\documentclass[runningheads]{llncs}
\usepackage{graphicx}

\usepackage{subfigure} 
\usepackage{amsmath}
\usepackage{amssymb}
\usepackage{mathtools}
\usepackage{caption} 
\begin{document}
\title{Two-Stream Deep Feature Modelling for Automated Video Endoscopy Data Analysis}
\titlerunning{Two-Stream Feature Modelling for Endoscopy Data Analysis}
%
\author{Harshala Gammulle \and Simon Denman \and Sridha Sridharan \and  Clinton Fookes}
\index{Gammulle, Harshala}
\index{Denman, Simon}
\index{Sridharan, Sridha}
\index{Fookes, Clinton}
\authorrunning{H. Gammulle et al.}
\institute{The Signal Processing, Artificial Intelligence and Vision Technologies (SAIVT), Queensland University of Technology, Australia. \\
\email{\{pranali.gammule, s.denman, s.sridharan, c.fookes\} @qut.edu.au}}


%
\maketitle              
\begin{abstract}

Automating the analysis of imagery of the Gastrointestinal (GI) tract captured during endoscopy procedures has substantial potential benefits for patients, as it can provide diagnostic support to medical practitioners and reduce mistakes via human error. To further the development of such methods, we propose a two-stream model for endoscopic image analysis. Our model fuses two streams of deep feature inputs by mapping their inherent relations through a novel relational network model, to better model symptoms and classify the image. In contrast to handcrafted feature-based models, our proposed network is able to learn features automatically and outperforms existing state-of-the-art methods on two public datasets: KVASIR and Nerthus. Our extensive evaluations illustrate the importance of having two streams of inputs instead of a single stream and also demonstrates the merits of the proposed relational network architecture to combine those streams. 

\end{abstract}
\section{Introduction}

In medicine, endoscopy procedures on the Gastrointestinal (GI) tract play an important role in supporting domain experts to track down abnormalities within the GI tract of a patient. Such abnormalities may be a symptom for a life-threatening disease such as colorectal cancer. This analysis is typically carried out manually by a medical expert, and detecting critical symptoms relies solely on the experience of the practitioner, and is susceptible to human error. As such, we seek to automate the process of endoscopic video analysis, providing support to human experts during diagnosis.        

Due to advancements in biomedical engineering, extensive research has been performed to support and improve the detection of anomalies via machine learning and computer vision techniques. These methods have shown great promise, and can detect abnormalities that can be easily missed by human experts \cite{guo2019triplemiccai,wang2019retinal,kumar2017kernel}. Yet automated methods face multiple challenges when analysing endoscopic videos, due to overlaps between symptoms and the difficult imaging conditions. 

Most previous endoscopy analysis approaches obtain a set of hand-crafted features and train models to detect abnormalities \cite{agrawal2017scl,liu2017hkbu}. For example, in \cite{liu2017hkbu} the encoded image features are obtained through a bidirectional marginal Fisher analysis (BMFA) and classified using a support vector machine (SVM). In \cite{naqvi2017ensemble}, local binary patterns (LBP) and edge histogram features are used with logistic regression. A limitation of these hand-crafted methods is that they are highly dependent on the domain knowledge of the human designer, and as such risk losing information that best describes the image. Therefore, through the advancement of deep learning approaches and due to their automatic feature learning ability, research has focused on deep learning methods. However, training these deep learning models from scratch is time consuming and requires a great amount of data. To overcome this challenge, transfer learning has been widely used; whereby a deep neural network that is trained on a different domain is adapted to the target domain through fine-tuning some or all layers. Such approaches have been widely used for anomaly detection in endoscopy videos obtained from the GI tract. The recent methods \cite{pogorelov2017kvasir,petscharnig2017inception} on computer aided video endoscopy analysis predominately extract discriminative features from a pre-trained convolutional neural network (CNN), and classify them using a classifier such as a Logistic Model Tree (LMT) or SVM. In \cite{borgli2019automatic}, a Bayesian optimisation method is used to optimise the hyper-parameters for a CNN based model for endoscopy data analysis. In \cite{agrawal2019evaluating}, the authors tested multiple existing pre-trained CNN network features to better detect abnormalities.

In \cite{lin2015bilinear} the authors propose an architecture that consists of two feature extractors. The outputs of these are multiplied using an outer product at each location of the image and are pooled to obtain an image descriptor. This architecture models local pairwise feature interactions. The authors of \cite{yu2018hierarchical} introduce a hierarchical bilinear pooling framework where they integrate multiple cross-layer bilinear modules to obtain information from intermediate convolution layers. In \cite{liu2016ssd} several skip connections between different layers were added to detect objects in different scales and aspect ratios. In contrast, the proposed work extracts semantic features from different CNN layers and explicitly models the relationship between these through a novel relation mapping network.
 
In this paper, we introduce a relational reasoning approach \cite{santoro2017simple} that is able to map the relationships among individual features extracted by a pre-trained deep neural network. We extract features from the mid layers of a pre-trained deep model and pass them through the relational network, which considers all possible relationships among individual features to classify an endoscopy image. Our primary evaluations are performed on the KVASIR dataset \cite{pogorelov2017kvasir}, containing endoscopic images and eight classes to detect. We also evaluate the proposed model on the Nerthus dataset \cite{pogorelov2017nerthus} to further demonstrate the effectiveness of the proposed model. For both datasets, the proposed method outperforms the existing state-of-the-art.

\begin{figure}[htbp]
        \centering
        	\includegraphics[width=1.0\linewidth]{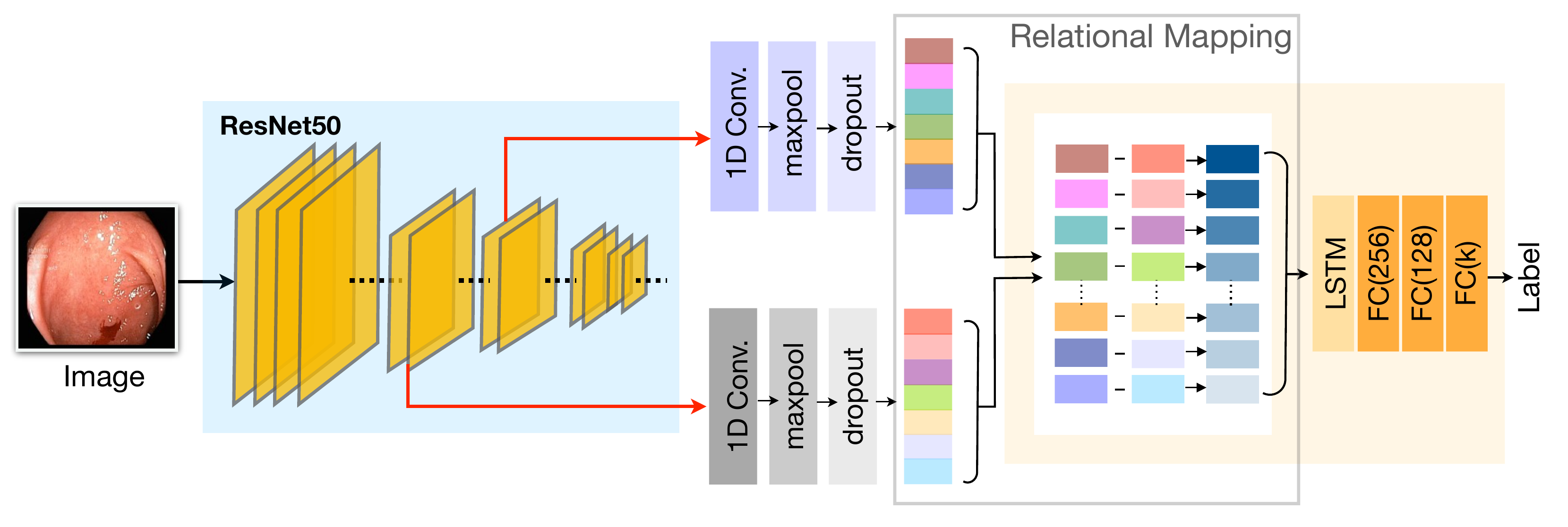}
	\caption{Proposed Model: The semantic features of the input image are extracted through two layers of a pre-trained ResNet50 network, and the relations among the encoded feature vectors are mapped through the relational network which facilitates the final classification.}
	\label{fig:model1}
\end{figure}

\section{Method}

In this paper, we propose a deep relational model that obtains deep information from two feature streams, that are combined to understand the class of the input endoscopy image. An overview of our proposed architecture is given in Figure \ref{fig:model1}.   

Training a CNN model from scratch is time consuming and requires a large dataset. Therefore, in practice it is more convenient to use a pre-trained network and adapt this to a target domain, and this has been shown to be an effective method in the computer vision \cite{gammulle2019forecasting,gammulle2019predicting} and medical domains \cite{agrawal2019evaluating,pogorelov2017kvasir}. To obtain the two feature streams we utilise a pre-trained ResNet50 \cite{resnet} network, trained on ImageNet \cite{imageNet}. Training on large-scale datasets such as ImageNet \cite{imageNet} improves the ability of the network to capture important patterns in input images and translate them into more discriminative feature vectors, that support different computer vision tasks. 

When extracting features from a pre-trained CNN model, features from earlier layers contain more local information than those from later layers; though later layers contain more semantic information \cite{gammulle2017two}. Thus combining such features offers more discriminative information and facilitates our final prediction task. In this study, we combine features from an earlier layer and a later layer from the pre-trained CNN model. This allows us to capture spatial and semantic features,  both of which are useful for accurate classification of endoscopy images. We avoid features from the final layers as they are over-compressed and do not contain information relating to our task, instead containing information primarily for the task the network is previously trained on (i.e. object detection). Our extracted features are further encoded through 1D convolutional and max pooling layers, and passed through a relational network to map the relationship between feature vectors, facilitating the final classification task.

\subsection{Semantic Feature Extractor }

The input image, $X$, is first passed through the Semantic Feature Extractor (SFE) module. The SFE is based on a ResNet50 pre-trained CNN, and features are extracted from two layers. We denote the respective features as,

\begin{equation}
\theta^{1} \in \mathbb{R}^{(W^{1}, H^{1}, D^{1})},
\end{equation}  
     
\begin{equation}
\theta^{2} \in \mathbb{R}^{(W^{2}, H^{2}, D^{2})},
\end{equation}

where $W^{1}, H^{1}, D^{1}$ and $W^{2}, H^{2}, D^{2}$ denote the sizes of the respective three-dimensional vectors. We reshape these vectors to two-dimensions such that they are of shape $(L^1 = W^{1} \times H^{1}, D^{1})$ and $(L^2 =W^{2} \times H^{2}, D^{2})$.

\subsection{Relational Network}

The resultant two-dimensional feature vectors are passed through separate 1D convolution functions, $f^{E^{1}}$ and $f^{E^{2}}$ , to further encode these features from the individual streams such that,
\begin{equation}
\beta^1 = f^{E^{1}}(\theta^{1}),
\end{equation}  
\begin{equation}
\beta^2 = f^{E^{2}}(\theta^{2}).
\end{equation}  

Then through a relational network, $f^{RN}$, we map all possible relations among the two input feature streams. Our relational network is inspired by the model introduced in \cite{santoro2017simple}. However, there exists a clear distinction between the proposed architecture and that of \cite{santoro2017simple}. \cite{santoro2017simple} utilises a relational network to map the relationships among the pixels in an input image. In the proposed work we illustrate that a relational network can be effectively utilised to map the correspondences among two distinct feature streams. We define the output of the relational network, $\gamma$, as, 
\begin{equation}
\gamma = f^{RN}([\beta^1, \beta^2]),
\end{equation}  
where $f^{RN}$ is composed of $f_g$ and $f_h$ which are Multi-Layer Perceptrons (MLPs), $i \in L^1$ and $j \in L^2$,
\begin{equation}
\gamma= f_h(\sum_{i}\sum_{j}f_g(\beta^1_i, \beta^2_j)).
\end{equation}

The resultant vector, $\gamma$, is passed through a decoding function, $f^D$, which is composed of a layer of LSTM cells \cite{hochreiter1997long}, and three fully connected layers to generate the classification of the input image,
\begin{equation}
y= f^D(\gamma).
\end{equation}

\section{Experiments}
\subsection{Datasets}

We utilise two publicly available endoscopy datasets, KVASIR and Nerthus, to demonstrate the capability of our model to analyse endoscopy images and detect varying conditions within the GI tract.

\textbf{The KVASIR Dataset \cite{pogorelov2017kvasir}} was released as part of the medical multimedia challenge presented by MediaEval \cite{riegler2017multimedia}. It is based on images obtained from the GI tract via an endoscopy procedure. The dataset is composed of images that are annotated and verified by medical doctors, and captures 8 different classes. The classes are based on three anatomical landmarks (z-line, pylorus, cecum), three pathological findings (esophagitis, polyps, ulcerative colitis) and two other classes (dyed and lifted polyps, dyed resection margins) related to the polyp removal process. Overall, the dataset contains 8,000 endoscopic images, with 1,000 image examples per class. We utilise the standard test set released by the dataset authors, where 4,000 samples are used for model training and 4,000 for testing.            

\textbf{The Nerthus Dataset \cite{pogorelov2017nerthus}} is composed of 2,552 images from 150 colonoscopy videos. The dataset contains 4 different classes defined by the Boston Bowel Preparation Scale (BBPS) score, that ranks the cleanliness of the bowel and is an essential part of a successful colonoscopy (the endoscopy examination of the bowel). The number of examples per class lies within the range 160 to 980, and the data is annotated by medical doctors. We use the training/testing splits provided by the dataset authors. 
 
\subsection{Metrics}

For the evaluations on the KVASIR dataset we utilise the metrics accuracy, precision, recall, F1-score, and matthews correlation coefficient (MCC) as suggested in \cite{pogorelov2017kvasir}. The evaluations on the Nerthus dataset utilise the accuracy metric.     

\subsection{Implementation Details}

We use a pre-trained ResNet50 \cite{resnet} network and extract features from two layers: `activation\_36' and `activation\_40'. Feature shapes are ($14\times14\times1024$) and ($14\times14\times256$) respectively. For the encoding of each feature stream we utilise a 1D convolution layer with a kernel size of 3 and 32 filters, followed by a BatchNorm\_ReLu \cite{Isola_CVPR2017} and a dropout layer, with a dropout rate of 0.25. The LSTM used has 300 hidden units and the output is further passed through three fully connected layers with the dimensionality of 256, 128 and k (number of classes) respectively. The model is trained using the RMSProp optimiser with a learning rate of 0.001 with a decay of $8\times10^{-9}$ for 100 epochs.  Implementation is completed in Keras \cite{keras} with a theano \cite{theano} backend.
     
\subsection{Results}

We use the KVASIR dataset for our primary evaluation and compare our results with recent state-of-the-art models (see Table \ref{tab:kvasir_results}). The first block of results in Table \ref{tab:kvasir_results} are the results obtained from various methods introduced for the MediaEval Challenge \cite{riegler2017multimedia} on the KVASIR data. In \cite{liu2017hkbu}, a dimensionality reduction method called bidirectional marginal Fisher analysis (BMFA) which uses a Support Vector Machine (SVM) is proposed; while in \cite{naqvi2017ensemble} a method that combines 6 different features (JCD, Edge Histogram, Color Layout, AutoColor Correlogram, LBP, Haralick) and uses a logistic regressor to classify these features is presented. Aside from hand-crafted feature based methods, in \cite{pogorelov2017kvasir} ResNet50 CNN features are extracted and fed to a Logistic Model Tree (LMT) classifier, and in \cite{petscharnig2017inception}, a GoogLeNet based model is employed.        
The authors in \cite{agrawal2017scl}, introduce an approach where they obtained a collection of hand-crafted features (Tamura, ColorLayout, EdgeHistogram and, AutoColorCorrelogram) and deep CNN network features (VGGNet and Inception-V3 features), and train a multi-class SVM. This model records the highest performance among the previous state-of-the-art methods. However, with two streams of deep feature fusion and relation mapping, our proposed model is able to outperform \cite{agrawal2017scl} by 2.3\% in accuracy, 5.1\% in precision, 4.5\% in recall, 5\% in F1-score, 5.1\% in MCC and 1.4\% in specificity. 

In \cite{agrawal2019evaluating}, the authors have tested extracting features from input endoscopic images through different pre-trained networks and classifying them through a multi-class SVM. In Table \ref{tab:kvasir_results} we show these results for ResNet50 features, MobileNet features and a combined deep feature obtained from multiple pre-trained CNN networks. In our proposed method, we also utilise features from a ResNet50 network, yet instead of naively combining features we utilise the proposed relational network to effectively attend to the feature vectors, and derive salient features for classification. 

\begin{table}[htbp]
\resizebox{1\textwidth}{!}{
\begin{tabular}{|l|c|c|c|c|c|c|}
\hline
Method            & Accuracy            & Precision           & Recall            & F1-score             & MCC            & Specificity           \\ \hline
Liu \cite{liu2017hkbu}               & 0.926          & 0.703          & 0.703          & 0.703          & 0.660          & 0.958           \\ \hline
Petsch \cite{petscharnig2017inception}            & 0.939          & 0.755          & 0.755          & 0.755          & 0.720          & 0.965          \\ \hline
Naqvi  \cite{naqvi2017ensemble}             & 0.942          & 0.767          & 0.774          & 0.767          & 0.736          & 0.966          \\ \hline
Pogorelov \cite{pogorelov2017kvasir}         & 0.957          & 0.829          & 0.826          & 0.826          & 0.802          & 0.975          \\ \hline
Agrawal   \cite{agrawal2017scl}        & 0.961          & 0.847          & 0.852          & 0.847          & 0.827          & 0.978          \\ \hline\hline \hline
ResNet50 \cite{agrawal2019evaluating}          & 0.611          & -              & -              & -              & -              & -              \\ \hline
MobileNet \cite{agrawal2019evaluating}         & 0.817          & -              & -              & -              & -              & -              \\ \hline
Combined feat. \cite{agrawal2019evaluating}    & 0.838          & -              & -              & -              & -              & -              \\ \hline \hline \hline
\textbf{Proposed} & \textbf{0.984} & \textbf{0.898} & \textbf{0.897} & \textbf{0.897} & \textbf{0.878} & \textbf{0.992} \\ \hline
\end{tabular}}
\caption{The evaluation results on KVASIR dataset.}
\label{tab:kvasir_results}
\end{table}

\begin{figure}[htbp]
        \centering
        	\includegraphics[width=.7\textwidth]{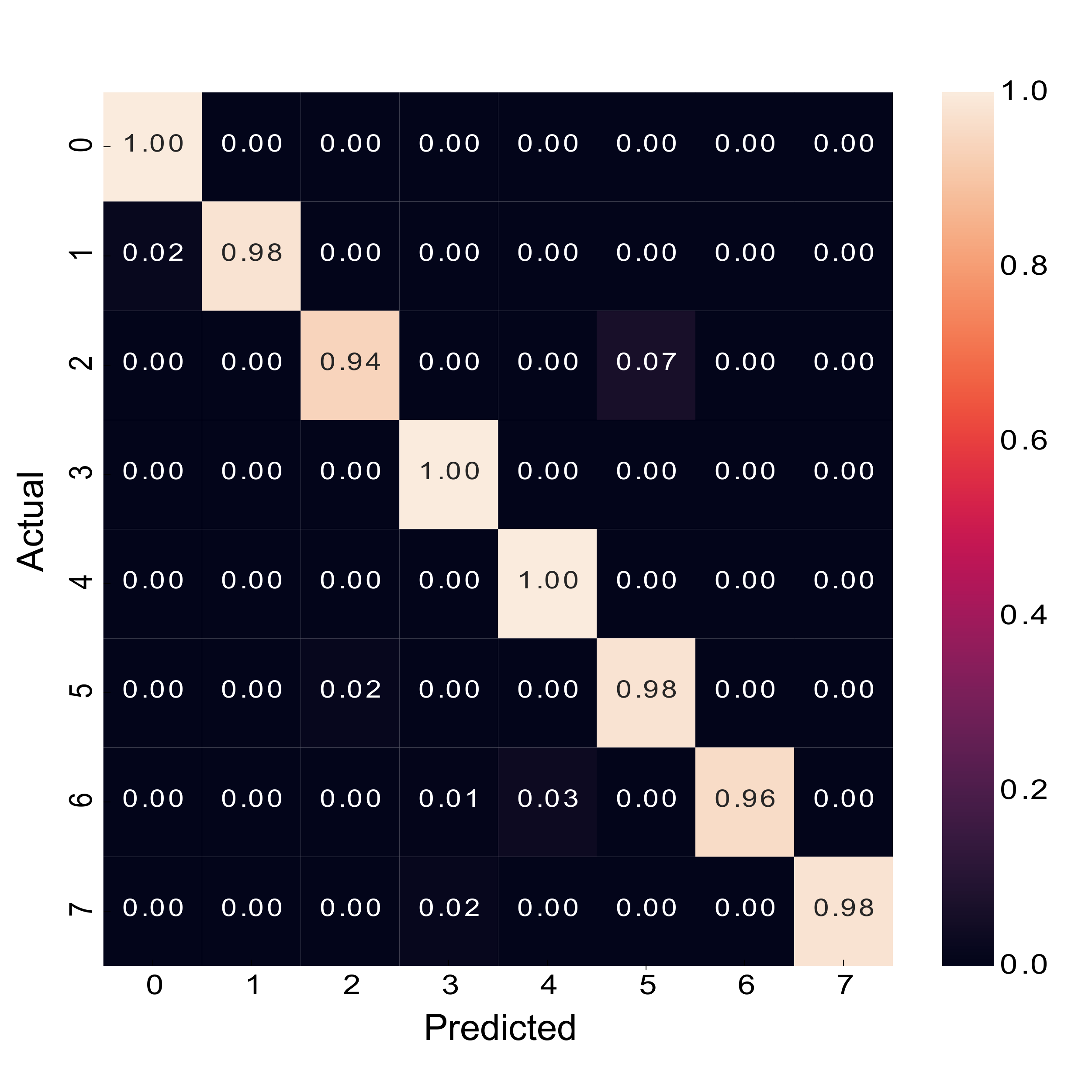}
	\caption{Confusion matrix illustration for the KVASIR dataset. }
	\label{fig:conf_mat}
\end{figure}

Figure \ref{fig:conf_mat} shows the confusion matrix for the evaluation results on the KVASIR dataset. For clarity we represent the classes as 0- `dyed-lifted-polyps’, 1- `dyed-resection-margins’, 2- `esophagitis’,  3- `normal-cecum’, 4- `normal-pylorus’, 5- `normal-z-line’, 6- `polyps’, 7- `ulcerative-colitis’. Confusions occur primarily between the normal-z-line and esophagitis classes, and a number of classes are classified correctly for all instances. 

\begin{figure}[!h]
       \centering
       \subfigure[Proposed two stream method]{\includegraphics[width=0.85\linewidth]{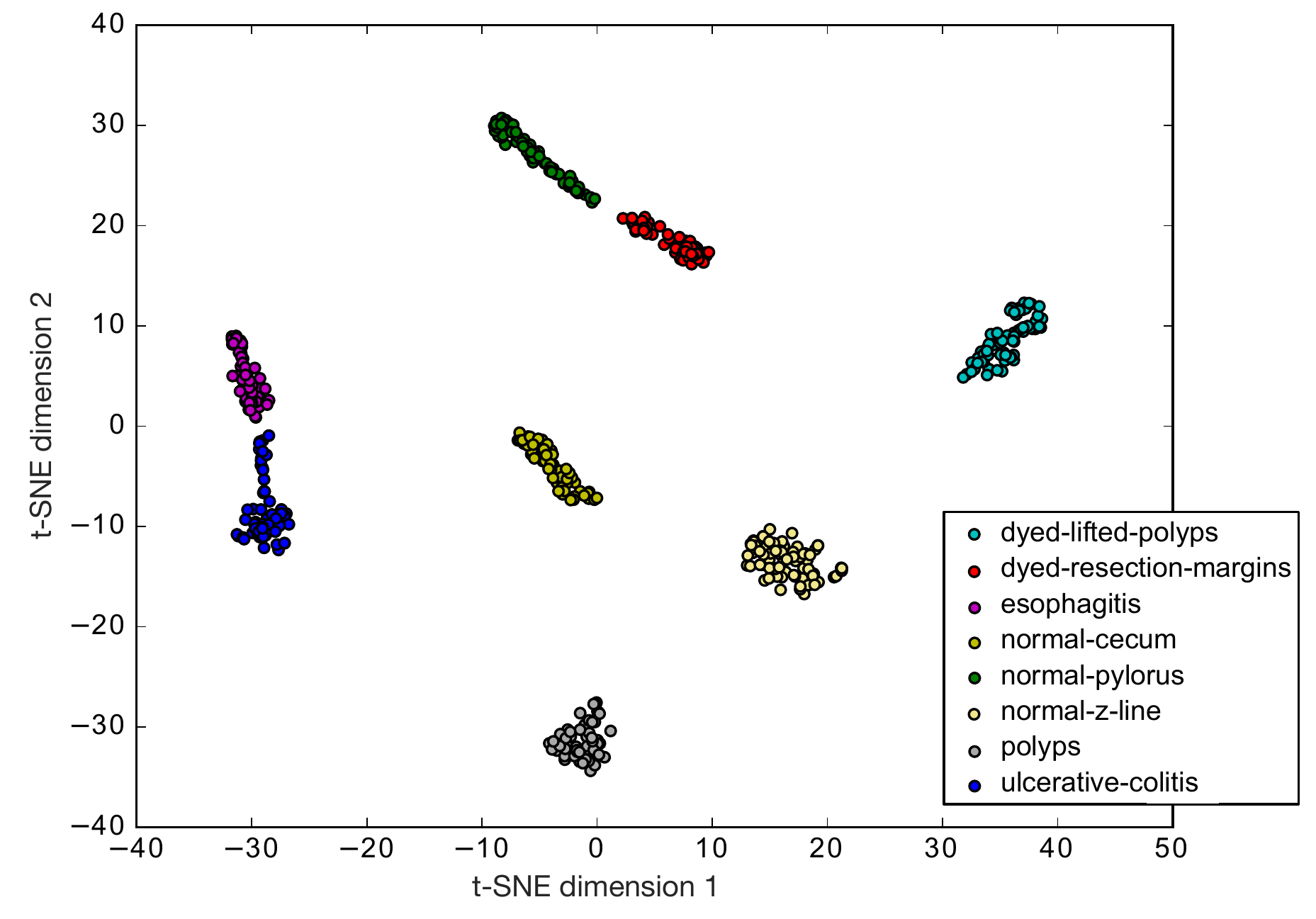}}
       \subfigure[Stream $\theta^1$]{\includegraphics[width=0.48\linewidth]{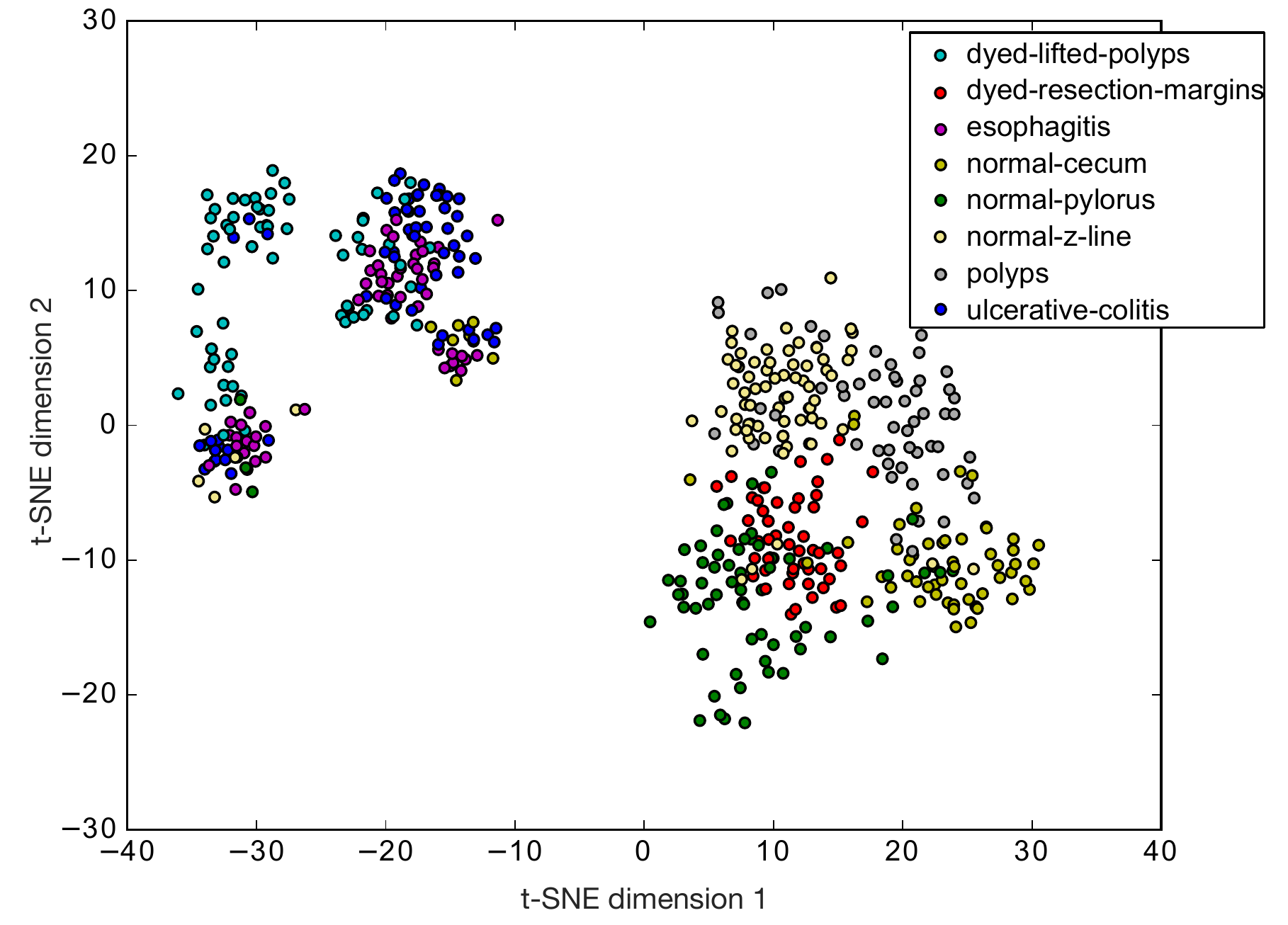}}  
       \subfigure[Stream $\theta^2$]{\includegraphics[width=0.48\linewidth]{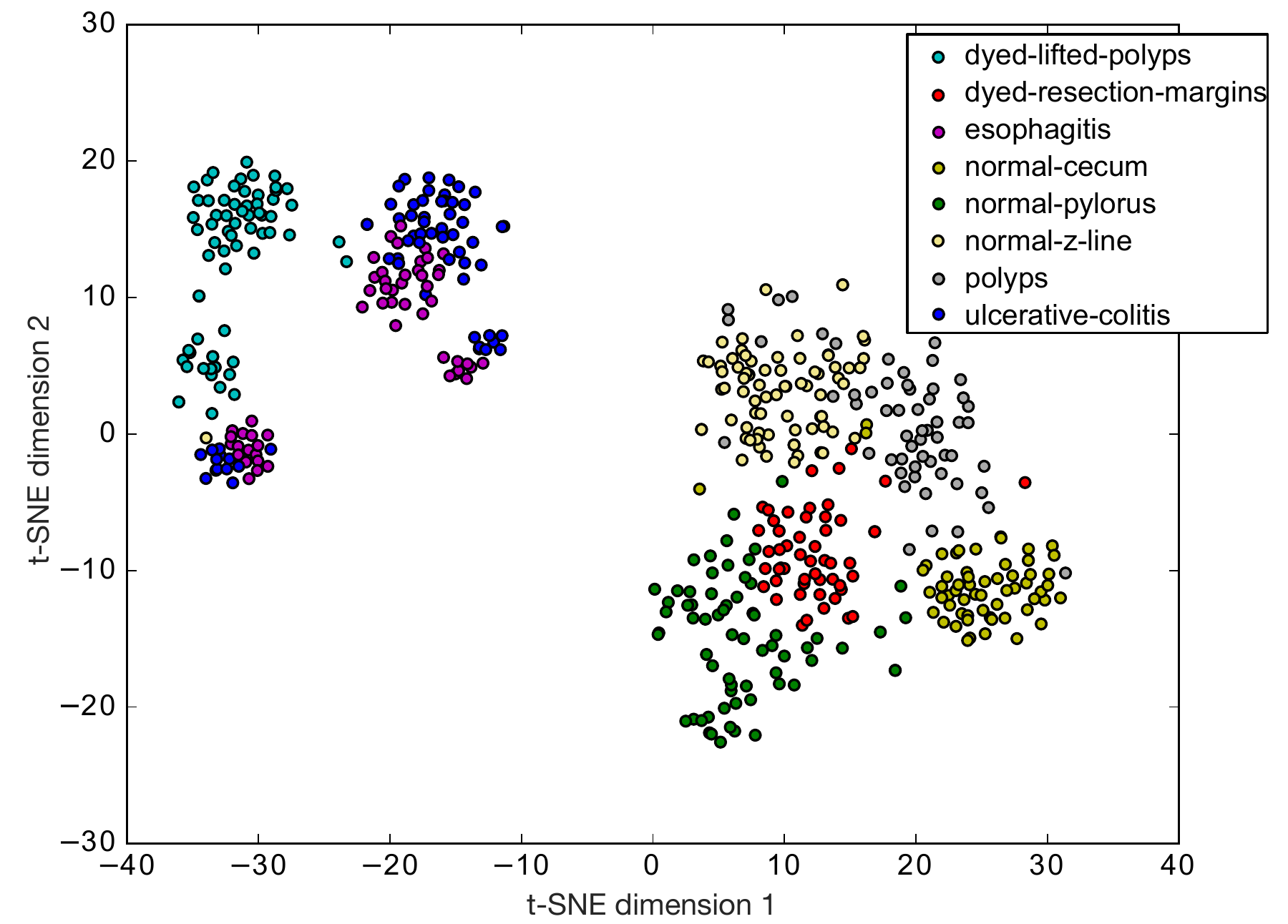}}  
	\caption{2D Visualisation of embeddings extracted from the LSTM layer of the proposed model (a) and two ablation models (b and c)}
	\label{fig:embeddings}
\end{figure}

To further illustrate the importance of our two-stream architecture and the value of the relational network for combining these feature streams, we visualise (in Figure \ref{fig:embeddings}) the activations obtained from the LSTM layer of the proposed model and two ablation models, each with only one input stream. The ablation model in Figure \ref{fig:embeddings} (b) receives the feature stream $\theta^1$ as the input, while ablation model in Figure \ref{fig:embeddings} (c) receives the feature stream $\theta^2$ as the input. In the ablation models (b) and (c), as in \cite{santoro2017simple} the relational network is used to model relationships within a single vector.

The activations are obtained for a randomly selected set of 500 images from the KVASIR test-set, and we use t-SNE \cite{maaten2008visualizing} to plot them in two dimensions.

Considering Fig. \ref{fig:embeddings} (a), we observe that samples from a particular class are tightly grouped and clear separation exists between classes. However, in the ablation models (b) and (c) we observe significant overlaps between the embeddings from different classes, indicating that the model is not capable of discriminating between those classes. These visualisations provide further evidence of the importance of utilising multiple input streams, and how they can be effectively fused together with the proposed relational model to learn discriminative features to support the classification task. 
    
To demonstrate the effectiveness of our model on different problem domains, we evaluated our model on the Nerthus dataset \cite{pogorelov2017nerthus}. While the task in this dataset, measuring the cleanliness of the bowel based on the BBPS value, is less challenging compared to the abnormality classification task in the KVASIR dataset, the Nerthus dataset provides a different evaluation scenario to investigate the generalisability of the proposed approach. We obtained a 100\% accuracy when predicting the BBPS value with our proposed model, while the baseline model of \cite{pogorelov2017nerthus} has only achieved a 95\% accuracy. This clearly illustrates the applicability of the proposed architecture for different classification tasks within the domain of automated endoscopy image analysis.
    
\section{Conclusion}

Endoscopy image analysis is a challenging task and automating this process can aid both the patient and the medical practitioner. Our approach is significantly different from the previous approaches that are based on obtaining handcrafted features or extracting pre-trained CNN features and learning a classifier based on these features. Our relational model, with two discriminative feature streams, is able to map dependencies between feature streams to help detect and identify salient features, and outperforms state-of-the-art methods for the KVASIR and Nerthus datasets. Furthermore, as our model learns the image to label mapping automatically, it is applicable for detecting abnormalities in other medical domains apart from the analysis of endoscopy images.      

\section{Acknowledgement}
The research presented in this paper was supported by an
Australian Research Council (ARC) grant DP170100632.

%
%

 \bibliographystyle{splncs04}
 \bibliography{miccai_1}

\begin{thebibliography}{10}
\providecommand{\url}[1]{\texttt{#1}}
\providecommand{\urlprefix}{URL }
\providecommand{\doi}[1]{https://doi.org/#1}

\bibitem{agrawal2019evaluating}
Agrawal, T., Gupta, R., Narayanan, S.: On evaluating cnn representations for
  low resource medical image classification. In: ICASSP 2019-2019 IEEE
  International Conference on Acoustics, Speech and Signal Processing (ICASSP).
  pp. 1363--1367. IEEE (2019)

\bibitem{agrawal2017scl}
Agrawal, T., Gupta, R., Sahu, S., Espy-Wilson, C.Y.: Scl-umd at the medico
  task-mediaeval 2017: Transfer learning based classification of medical
  images. In: MediaEval (2017)

\bibitem{theano}
Al-Rfou, R., Alain, G., Almahairi, A., Angermueller, C., Bahdanau, D., Ballas,
  N., Bastien, F., Bayer, J., Belikov, A., Belopolsky, A., et~al.: Theano: A
  python framework for fast computation of mathematical expressions. arXiv
  preprint arXiv:1605.02688  \textbf{472}, ~473 (2016)

\bibitem{borgli2019automatic}
Borgli, R.J., Stensland, H.K., Riegler, M.A., Halvorsen, P.: Automatic
  hyperparameter optimization for transfer learning on medical image datasets
  using bayesian optimization. In: 2019 13th International Symposium on Medical
  Information and Communication Technology (ISMICT). pp.~1--6. IEEE (2019)

\bibitem{keras}
Chollet, F., et~al.: Keras. \url{https://keras.io} (2015)

\bibitem{gammulle2017two}
Gammulle, H., Denman, S., Sridharan, S., Fookes, C.: Two stream lstm: A deep
  fusion framework for human action recognition. In: Applications of Computer
  Vision (WACV), 2017 IEEE Winter Conference on. pp. 177--186. IEEE (2017)

\bibitem{gammulle2019forecasting}
Gammulle, H., Denman, S., Sridharan, S., Fookes, C.: Forecasting future action
  sequences with neural memory networks. British Machine Vision Conference
  (BMVC)  (2019)

\bibitem{gammulle2019predicting}
Gammulle, H., Denman, S., Sridharan, S., Fookes, C.: Predicting the future: A
  jointly learnt model for action anticipation. In: Proceedings of the IEEE
  International Conference on Computer Vision. pp. 5562--5571 (2019)

\bibitem{guo2019triplemiccai}
Guo, X., Yuan, Y.: Triple anet: Adaptive abnormal-aware attention network for
  wce image classification. In: International Conference on Medical Image
  Computing and Computer-Assisted Intervention (MICCAI). pp. 293--301. Springer
  (2019)

\bibitem{resnet}
He, K., Zhang, X., Ren, S., Sun, J.: Deep residual learning for image
  recognition. In: Proceedings of the IEEE Conference on Computer Vision and
  Pattern Recognition. pp. 770--778 (2016)

\bibitem{hochreiter1997long}
Hochreiter, S., Schmidhuber, J.: Long short-term memory. Neural computation
  \textbf{9}(8),  1735--1780 (1997)

\bibitem{Isola_CVPR2017}
Isola, P., Zhu, J.Y., Zhou, T., Efros, A.A.: Image-to-image translation with
  conditional adversarial networks. In: The IEEE Conference on Computer Vision
  and Pattern Recognition (CVPR) (July 2017)

\bibitem{kumar2017kernel}
Kumar, N., Rajwade, A.V., Chandran, S., Awate, S.P.: Kernel
  generalized-gaussian mixture model for robust abnormality detection. In:
  International Conference on Medical Image Computing and Computer-Assisted
  Intervention (MICCAI). pp. 21--29. Springer (2017)

\bibitem{lin2015bilinear}
Lin, T.Y., RoyChowdhury, A., Maji, S.: Bilinear cnn models for fine-grained
  visual recognition. In: Proceedings of the IEEE international conference on
  computer vision. pp. 1449--1457 (2015)

\bibitem{liu2016ssd}
Liu, W., Anguelov, D., Erhan, D., Szegedy, C., Reed, S., Fu, C.Y., Berg, A.C.:
  Ssd: Single shot multibox detector. In: European conference on computer
  vision. pp. 21--37. Springer (2016)

\bibitem{liu2017hkbu}
Liu, Y., Gu, Z., Cheung, W.K.: Hkbu at mediaeval 2017 medico: Medical
  multimedia task  (2017)

\bibitem{maaten2008visualizing}
Maaten, L.v.d., Hinton, G.: Visualizing data using t-sne. Journal of machine
  learning research  \textbf{9}(Nov),  2579--2605 (2008)

\bibitem{naqvi2017ensemble}
Naqvi, S.S.A., Nadeem, S., Zaid, M., Tahir, M.A.: Ensemble of texture features
  for finding abnormalities in the gastro-intestinal tract. In: MediaEval
  (2017)

\bibitem{petscharnig2017inception}
Petscharnig, S., Sch{\"o}ffmann, K., Lux, M.: An inception-like cnn
  architecture for gi disease and anatomical landmark classification. In:
  MediaEval (2017)

\bibitem{pogorelov2017kvasir}
Pogorelov, K., Randel, K.R., Griwodz, C., Eskeland, S.L., de~Lange, T.,
  Johansen, D., Spampinato, C., Dang-Nguyen, D.T., Lux, M., Schmidt, P.T.,
  et~al.: Kvasir: A multi-class image dataset for computer aided
  gastrointestinal disease detection. In: Proceedings of the 8th ACM on
  Multimedia Systems Conference. pp. 164--169 (2017)

\bibitem{pogorelov2017nerthus}
Pogorelov, K., Randel, K.R., de~Lange, T., Eskeland, S.L., Griwodz, C.,
  Johansen, D., Spampinato, C., Taschwer, M., Lux, M., Schmidt, P.T., et~al.:
  Nerthus: A bowel preparation quality video dataset. In: Proceedings of the
  8th ACM on Multimedia Systems Conference. pp. 170--174 (2017)

\bibitem{riegler2017multimedia}
Riegler, M., Pogorelov, K., Halvorsen, P., Griwodz, C., Lange, T., Randel, K.,
  Eskeland, S., Nguyen, D., Tien, D., Lux, M., et~al.: Multimedia for medicine:
  the medico task at mediaeval 2017  (2017)

\bibitem{imageNet}
Russakovsky, O., Deng, J., Su, H., Krause, J., Satheesh, S., Ma, S., Huang, Z.,
  Karpathy, A., Khosla, A., Bernstein, M., Berg, A.C., Fei-Fei, L.: {ImageNet
  Large Scale Visual Recognition Challenge}. International Journal of Computer
  Vision (IJCV)  \textbf{115}(3),  211--252 (2015)

\bibitem{santoro2017simple}
Santoro, A., Raposo, D., Barrett, D.G., Malinowski, M., Pascanu, R., Battaglia,
  P., Lillicrap, T.: A simple neural network module for relational reasoning.
  In: Advances in neural information processing systems. pp. 4967--4976 (2017)

\bibitem{wang2019retinal}
Wang, X., Ju, L., Zhao, X., Ge, Z.: Retinal abnormalities recognition using
  regional multitask learning. In: International Conference on Medical Image
  Computing and Computer-Assisted Intervention (MICCAI). pp. 30--38. Springer
  (2019)

\bibitem{yu2018hierarchical}
Yu, C., Zhao, X., Zheng, Q., Zhang, P., You, X.: Hierarchical bilinear pooling
  for fine-grained visual recognition. In: Proceedings of the European
  conference on computer vision (ECCV). pp. 574--589 (2018)

\end{thebibliography}

\end{document}